\documentclass[10pt,journal]{IEEEtran}
\usepackage{enumitem,calc}
\usepackage{graphicx}
\usepackage{subfigure}
\usepackage{amsmath}
\usepackage{array}
\usepackage{subfig}
\usepackage{tabu}
\usepackage{amssymb}
\usepackage{multirow}
\usepackage{booktabs}
\usepackage{graphicx}

\usepackage{cite}
\usepackage{xcolor}
\usepackage{amsfonts}
\usepackage{makecell}
\usepackage[noend]{algpseudocode}
\usepackage{soul}
\soulregister\cite7
\soulregister\ref7
\soulregister\eqref7
\ifCLASSINFOpdf
\else
\fi
\begin{document}
%
\title{ G$^2$HFNet: GeoGran-Aware Hierarchical Feature Fusion Network for Salient Object Detection in Optical Remote Sensing Images}

\author{Bin~Wan,
	Runmin~Cong,
	Xiaofei~Zhou,
	Hao~Fang,
	Chengtao~Lv,
	and Sam~Kwong, \emph{Fellow}, \emph{IEEE}
		\thanks{This work was supported in part by  the opening project of State Key Laboratory of Autonomous Intelligent Unmanned Systems under Grant ZZKF2025-2-8, in part by the National Natural Science Foundation of China under Grant 62471278 and 62271180, in part by the Hong Kong GRF-RGC General Research Fund under Grant 13200425, and in part by the Research Grants Council of the Hong Kong Special Administrative Region, China under Grant STG5/E-103/24-R. \emph{(Corresponding author: Runmin Cong)}.}
	\thanks{ Bin Wan, Runmin Cong and Hao Fang are with the School of Control Science and Engineering, Shandong University, Jinan 250061, China, and also with State Key Laboratory of Autonomous Intelligent Unmanned Systems, Shanghai 201210, China. (E-mail:  wanbinxueshu@icloud.com; rmcong@sdu.edu.cn; fanghaook@mail.sdu.edu.cn).}
	
	\thanks{ Xiaofei Zhou is with School of Automation, Hangzhou Dianzi University, Hangzhou 310018, China. (E-mail:  zxforchid@outlook.com).}
	\thanks{Chengtao lv is with School of information engineering, Huzhou University, Huzhou 313000, China. (E-mail: chengtaolv@outlook.com).}
	
	\thanks{ Sam Kwong is with the School of Data Science, Lingnan University, Tuen
		Mun, Hong Kong. (E-mail: samkwong@ln.edu.hk).}
	
		
	
}

\maketitle

\begin{abstract}
	Remote sensing images captured from aerial perspectives often exhibit significant scale variations and complex backgrounds, posing challenges for salient object detection (SOD). Existing methods typically extract multi-level features at a single scale using uniform attention mechanisms, leading to suboptimal representations and incomplete detection results. To address these issues, we propose a GeoGran-Aware Hierarchical Feature Fusion Network (G$^2$HFNet) that fully exploits geometric and granular cues in optical remote sensing images. Specifically, G$^2$HFNet adopts Swin Transformer as the backbone to extract multi-level features and integrates three key modules: the multi-scale detail enhancement (MDE) module to handle object scale variations and enrich fine details, the dual-branch geo-gran complementary (DGC) module to jointly capture fine-grained details and positional information in mid-level features, and the deep semantic perception (DSP) module to refine high-level positional cues via self-attention. Additionally, a local-global guidance fusion (LGF) module is introduced to replace traditional convolutions for effective multi-level feature integration. Extensive experiments demonstrate that G$^2$HFNet achieves high-quality saliency maps and significantly improves detection performance in challenging remote sensing scenarios.


\end{abstract}



\IEEEpeerreviewmaketitle

\section{Introduction}
\IEEEPARstart{A}{s} a novel computer vision task, salient object detection (SOD)  was developed to mimic the human ability to rapidly identify key object information within an image. Unlike object detection tasks that involve drawing bounding boxes around targets, SOD is a pixel-level binary classification task that requires precise mining and extraction of various object-related information. Therefore, SOD has been widely applied to a variety of visual scene tasks\cite{jing2021occlusion,xing2025towards,cong2025breaking,cong2025trnet}, such as medical image segmentation \cite{cong2022bcs,zhou2024uncertainty,wu2024cnn}, video image segmentation \cite{chen2022novel,su2023unified,qin2025sight,fang2025decoupled}, light field object detection \cite{chen2023light,gao2023thorough}, semantic segmentation \cite{an2023dual,lian2024diving,zheng2024deep,cong2024query,cong2025divide,chen2025empowering,cong2025uis,chen2025replay} and multi-modal segmentation \cite{gao2024highly,han2025perceptual,xiong2025mm}.

In the early stage, salient object detection mainly focused on processing natural scene images and experienced explosive growth alongside the rapid development of deep learning technologies \cite{li2016single}. For example, Wang \emph{et al}. \cite{wang2023pixels} proposed a  multiple enhancement network  with a multi-level hybrid loss and a flexible multiscale feature enhancement module to improve accuracy and robustness in complex scenes. In \cite{liu2022poolnet+}, Liu \emph{et al}.  explored pooling techniques for salient object detection by proposing a global guidance module and a feature aggregation module to progressively refine semantic features and produce detail-enriched saliency maps. With the growing demand for high-precision analysis in aerial and satellite imagery, salient object detection has gradually expanded to remote sensing images, where objects often vary significantly in scale and orientation, exhibit complex backgrounds, and present lower contrast, posing new challenges and research opportunities. However, directly applying methods designed for natural images to remote sensing scenarios often results in a sharp decline in performance. Consequently, many researchers are dedicated to developing efficient detection networks specifically tailored for remote sensing images to mitigate the negative impacts of varying object scales and orientations \cite{han2025orsidiff,xing2025lightweight}, as well as overwhelming background context redundancy. For instance, Cong \emph{et al}. \cite{cong2021rrnet} proposed a relational reasoning network that leverages parallel multiscale attention to address complex backgrounds and scale variations in optical remote sensing images. Besides, Wang \emph{et al}. \cite{wang2022multiscale} designed a multiscale detection network to address challenges like insufficient feature extraction, tiny object detection, and background interference by integrating multiple object information.

\begin{figure}[t]
	\centering
	\includegraphics[width=0.45\textwidth]{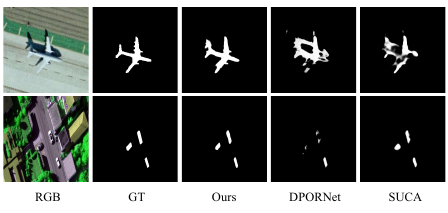}
	\caption{\small{Remote-sensing objects at different scales.}}
	\label{fig_ker}
\end{figure}

However, since remote sensing images are captured from a top-down aerial perspective\cite{cong2025generalized}, the  objects often exhibit significant scale variations, bringing significant challenges to detection methods,  as shown in Fig. \ref{fig_ker}. To capture fine details, most existing methods extract and refine multi-level features at a single scale using attention mechanisms. This strategy often fails to effectively handle large-scale variations and complex contextual information, leading to suboptimal feature representation and incomplete saliency detection results. Although Zhou \emph{et al}. \cite{zhou2022edge} introduced a branch that compresses image scales to cope with scale variations, this approach inevitably results in information loss. Besides, the multi-level features obtained from the encoder often contain different types of object information ( \emph{i.e.}, high-level features focus on positional information, while low-level features emphasize fine details). However, most existing methods adopt the same optimization strategy for all feature levels, which fails to effectively exploit object-specific information. To address this issue, Li  \emph{et al}. \cite{li2023lightweight} designed three specialized modules to extract different types of features, employing mutual attention on mid-level features to capture contextual information. However, they overlooked the fact that mid-level features also contain valuable detail and positional information.


\begin{figure*}[t]
\centering
\includegraphics[width=0.8\textwidth]{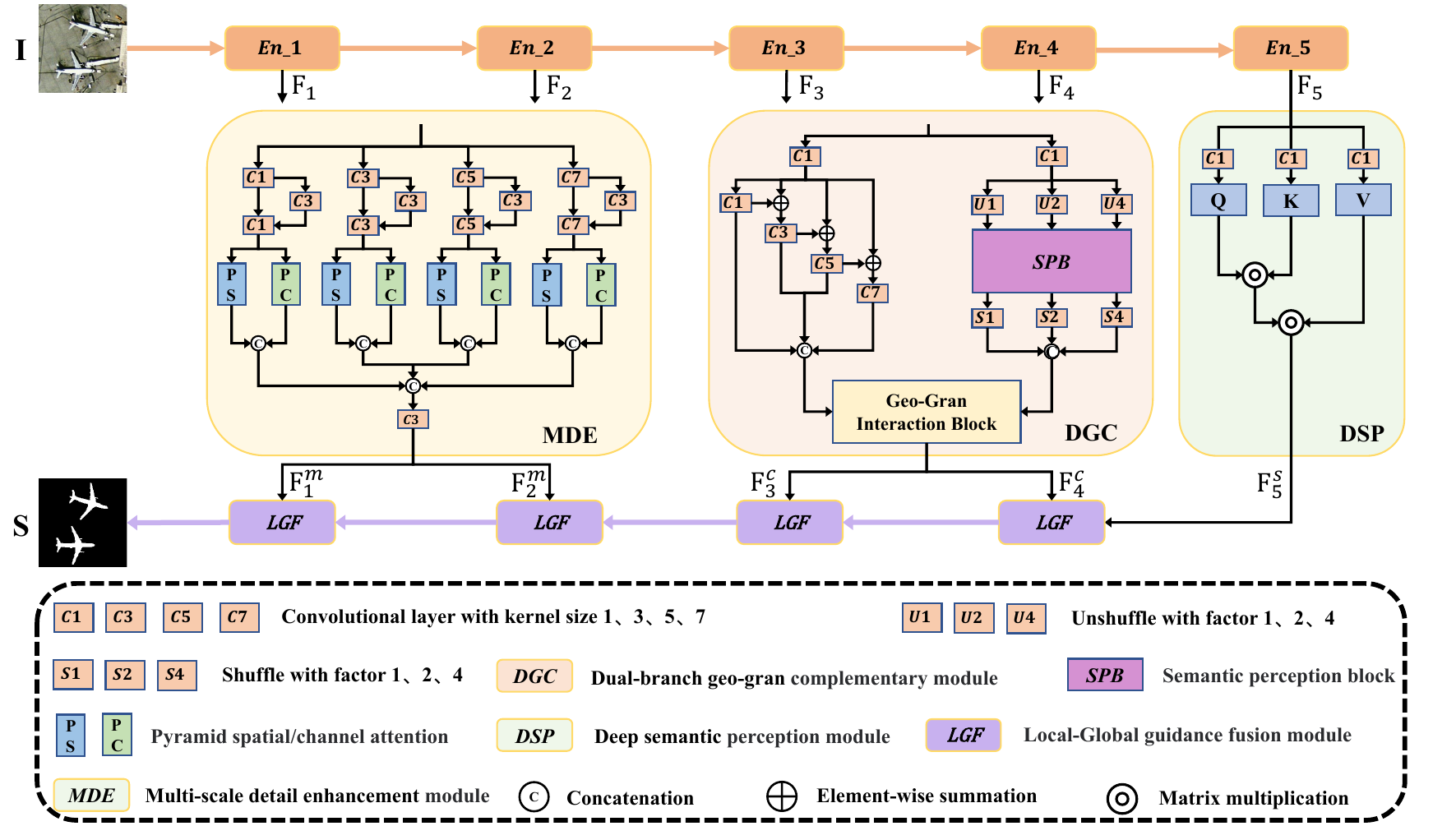}
\caption{\small{The overall architecture of proposed G$^2$HFNet.}}
\label{fig_ov}
\end{figure*}

To overcome the limitations mentioned above, in this paper, we propose a GeoGran-Aware Hierarchical Feature Fusion Network (G$^2$HFNet) for Salient Object Detection which fully exploits the geometric and granular cues in optical remote sensing images. As shown in Fig. \ref{fig_ov}, our G$^2$HFNet adopts SwinTransformer as the backbone to extract multi-level features, and designs  the multi-scale detail enhancement (MDE) module, dual-branch geo-gran complementary (DGC) module and deep semantic perception (DSP) module to extract and optimize different types of features. Besides, the local-global guidance fusion (LGF) module is proposed to is proposed to replace traditional 3$\times$3 convolutions for multi-level feature fusion. 
To be specific, first, to handle the pronounced scale variations commonly seen in optical remote sensing images (where targets may appear extremely small, elongated, or dispersed due to high-altitude imaging), and to extract fine structural details, the multi-scale detail enhancement (MDE) module is integrated into the low-level feature layers. Different from previous methods that apply convolutions with varying kernel sizes directly to single-scale features, the MDE module adopts a U-shaped structure across different convolutional layers, enabling features to undergo explicit scale transformation and thus capture richer fine-grained details across diverse object sizes. This multi-scale encoding-decoding pathway allows the network to better perceive small targets while preserving structural integrity for large or irregularly shaped objects. Subsequently, to further optimize these scale-aware features, we introduce the pyramid spatial and channel attention blocks, which deform the traditional spatial and channel attention mechanisms by performing multi-scale sampling along the spatial and channel dimensions, respectively. This design enables the network to selectively emphasize informative regions and semantic responses at multiple scales, enhancing its ability to cope with the heterogeneous object sizes and complex spatial distributions inherent in remote sensing scenes.
For the mid-level features, which inherently contain both rich fine-grained details and essential positional cues, we design a dual-branch geo-gran complementary (DGC) module comprising a granular branch and a geometric branch. The granular branch progressively captures multi-scale detailed information by cascading convolutional layers with different kernel sizes, enabling the network to better model the diverse textures and structural variations present in remote sensing targets. Meanwhile, the geometric branch performs multi-scale feature sampling and enhances positional cues through a semantic perception block, thereby improving the model's ability to cope with the irregular and widely distributed object locations characteristic of remote sensing imagery. After that, the geometric and granular features are integrated in the geometric-granular interaction block, which allows the network to adaptively fuse these complementary detail and positional components into a unified mid-level representation.
In addition, since the high-level features extracted by the encoder provide relatively reliable positional cues, we apply a self-attention mechanism exclusively to these features to further enhance their spatial precision. This is especially crucial in optical remote sensing images, where objects often appear at arbitrary and unpredictable locations due to wide-area coverage and varying imaging viewpoints. By modeling long-range dependencies through self-attention, the network can better capture global spatial relationships and maintain robust positional awareness even when targets are irregularly distributed across the scene.
Finally, to address the insufficient multi-level feature fusion in conventional U-Net-based decoders-where a single convolution provides limited guidance-we design the local-global guidance fusion (LGF) module, which is specifically introduced to fully integrate the detailed and positional information generated by the MDE, DGC, and DSP modules. LGF includes a local guidance stage that strengthens fine-scale structural details through gated convolution and a global guidance stage that uses high-level cues to direct low-level features toward target-relevant regions, thereby allowing  detail information and positional context to interact in a complementary manner. With this coordinated process, LGF achieves more complete integration of multi-level cues and produces higher-quality detection maps.


The main contributions of this paper are summarized as follows,
\begin{enumerate}[leftmargin=*]
\item We propose a novel GeoGran-Aware Hierarchical Feature Fusion Network (G$^2$HFNet) for Salient Object Detection in Optical Remote Sensing  Images which consisits the multi-scale detail enhancement (MDE) module, dual-branch geo-gran complementary (DGC) module, deep semantic perception (DSP) module and local-global guidance fusion (LGF) module. Experimental results on  three public remote sensing  image dataset prove that our G$^2$HFNet outperforms the other state-of-the-art  detection methods.
\item  We propose the multi-scale detail enhancement (MDE) module which is integrated into low-level features with a U-Net structure, captures richer details after scale transformation, while the pyramid spatial and channel attention blocks further enhance features via multi-scale spatial and channel sampling.
\item We propose a dual-branch geo-gran complementary (DGC) module, where the granular branch captures fine details and the geometric branch enhances positional information. 
\item We design deep semantic perception (DSP) module and local-global guidance fusion (LGF) module to  optimize the location cues and integrate multi-level features along both local and global dimension.
\end{enumerate}

\section{Related Works}
\subsection{Salient Object Detection for Natural Scene Images}
In the early stages of salient object detection development, natural images were widely utilized in this field, leading to the emergence of various efficient detection methods. For example, in \cite{zhuge2022salient}, Zhu \emph{et al}. proposed an integrity cognition network that integrates diverse feature aggregation, channel enhancement, and part-whole verification to achieve more comprehensive and coherent detection. In \cite{wang2023pixels}, Wang \emph{et al}. proposed a  multiple enhancement network, which integrates multi-level hybrid loss, a flexible multiscale feature enhancement module, and an iterative training strategy to achieve comprehensive and refined saliency detection. In \cite{liu2023tcgnet}, Liu \emph{et al}. proposed a type-correlation guidance network, which leverages multi-type cue correlation and type interaction attention to jointly enhance saliency prediction. In \cite{yang2024saliency}, Yang \emph{et al}. proposed an end-to-end set prediction framework that jointly integrates saliency and edge features through random edge neighborhood sampling, enabling simultaneous extraction and prediction of both edge and salient object maps without the need for complex multi-branch designs or repeated training. In \cite{zhu2025dc}, Zhu \emph{et al}. introduced the divide-and-conquer strategy into salient object detection by guiding multiple encoders to solve different subtasks and aggregating their semantic features via a decoder equipped with two-level residual nested-aspp modules, effectively addressing sparse receptive fields and limitations of conventional U-shaped structures.
\subsection{Salient Object Detection for ORSI}
Considering the remote sensing images play a vital role in urban planning, land use monitoring, environmental protection, and disaster assessment, significantly improving efficiency and supporting rapid decision-making, the salient object detection based on remote sensing images has rapidly advanced. In the early stage, Zhang \cite{zhang2020dense} and Tu \cite{tu2021orsi} proposed two datasets, EORSSD and ORSI-4199, specifically designed to improve detection performance in remote sensing imagery. After that, researchers have conducted continuous studies on salient object detection in remote sensing images based on these datasets. For example, in \cite{wang2022hybrid}, Wang \emph{et al}. integrated a hybrid encoder, gated fold-aspp, adjacent feature alignment to address the cluttered backgrounds, scale variations, and complex edges in the optical remote sensing images. In \cite{li2023salient}, Li \emph{et al}. proposed  a global-to-local framework to effectively addresses the limitations of CNN-based methods in capturing global context, enhances orientation adaptability, and improves the accuracy of salient object localization in optical remote sensing images. In \cite{huang2024exploiting}, Huang \emph{et al}. designed a novel memory-based context propagation network for RSI-SOD, which integrates a cross-image dual memory module to leverage dataset-level contextual information and a shared attention-guided fusion module to address the limitation of relying solely on individual image context. In \cite{gu2024prnet}, Gu \emph{et al}. proposed a parallel refinement network with a group feature learning  framework for ORSI-SOD, which introduces a parallel refinement module composed of three identical blocks to address the imbalance between performance and efficiency. In \cite{gu2025optical}, Gu \emph{et al}. integrates a bidirectional cross-attention module to enhance cross-layer semantic and detail representations, and an attention restoration module to address attention vacuity through foreground-background decoupling and local supplementation. In \cite{ge2025semantic}, Ge \emph{et al}. proposed SAANet, which enhances global semantic perception through the multi-scale object extraction mamba module and achieves accurate salient region detection and redundancy suppression via the adaptive feature fusion module by exploring cross-level feature correlations to address inherent complexity of remote sensing imagery. In \cite{ai2025lightweight}, Ai \emph{et al}. introduced a lightweight framework with MobileNetV2 as the encoder, integrating multiple modules for multiscale feature enhancement and employing a multistream progressive decoding strategy to refine saliency, edge, and skeleton maps for improved boundary accuracy and robustness. In \cite{li2025adversarial}, Li \emph{et al}. introduced a new defense framework that enhances adversarial robustness by implicitly enriching feature representations through a two-stage noise search-optimization strategy and two global-guided modules, which collaboratively strengthen low-level texture cues and high-level semantic comprehension, thereby mitigating adversarial perturbations and improving detection reliability.
\section{Proposed Framework}
\subsection{Overview of Proposed G$^2$HFNet}
In this paper, we propose a GeoGran-Aware Hierarchical Feature Fusion Network for Salient Object Detection in Optical Remote Sensing Images which comprises four key components (\emph{i.e.}, multi-scale detail enhancement (MDE) module, dual-branch geo-gran complementary (DGC) module, deep semantic perception (DSP) module and local-global guidance fusion (LGF) module). To be specific, as shown in Fig. \ref{fig_ov}, the input image $\mathbf{I}$ with $4\times3\times384\times384$ resolution is first fed into the backbone network to extracti multi-level features $\{\mathbf{F}_i\}_{i=1}^5$. After that, to extract and optimize the detail information embedded in the low-level featues $\{\mathbf{F}_i\}_{i=1}^2$, the MDE module employs multiple U-Net structure branches with convolutions of varying kernel sizes to extract fine object details $\{\mathbf{F}_{1/2}^j\}_{j=1}^4$ under different receptive fields, and pyramid spatial and channel attention is applied to each feature to optimize information at multiple scales, the refined features $\{\mathbf{F}_{1/2}^{p_j}\}_{j=1}^4$ are then concatenated and fused to generate the final detailed feature representation $\mathbf{F}_{1/2}^m$. For the mid-level features $\{\mathbf{F}_i\}_{i=3}^4$, the DGC module captures detailed feature $\mathbf{F}_{3/4}^d$  and positional feature $\mathbf{F}_{3/4}^s$  through granular branch and geometric branch, respectively, and then fuses them via an interaction module to generate features $\mathbf{F}_{3/4}^c$ enriched with comprehensive object information. Besides, for the high-level feature $\mathbf{F}_5$, the DSP module adopt the self-attention mechanism to optimize location cues to obtain feature $\mathbf{F}_5^s$. Finally, multi-level features are progressively fused through the LGF module to generate high-quality detection results $\mathbf{S}$.

\begin{figure}[!t]
	\centering
	\includegraphics[width=0.49\textwidth]{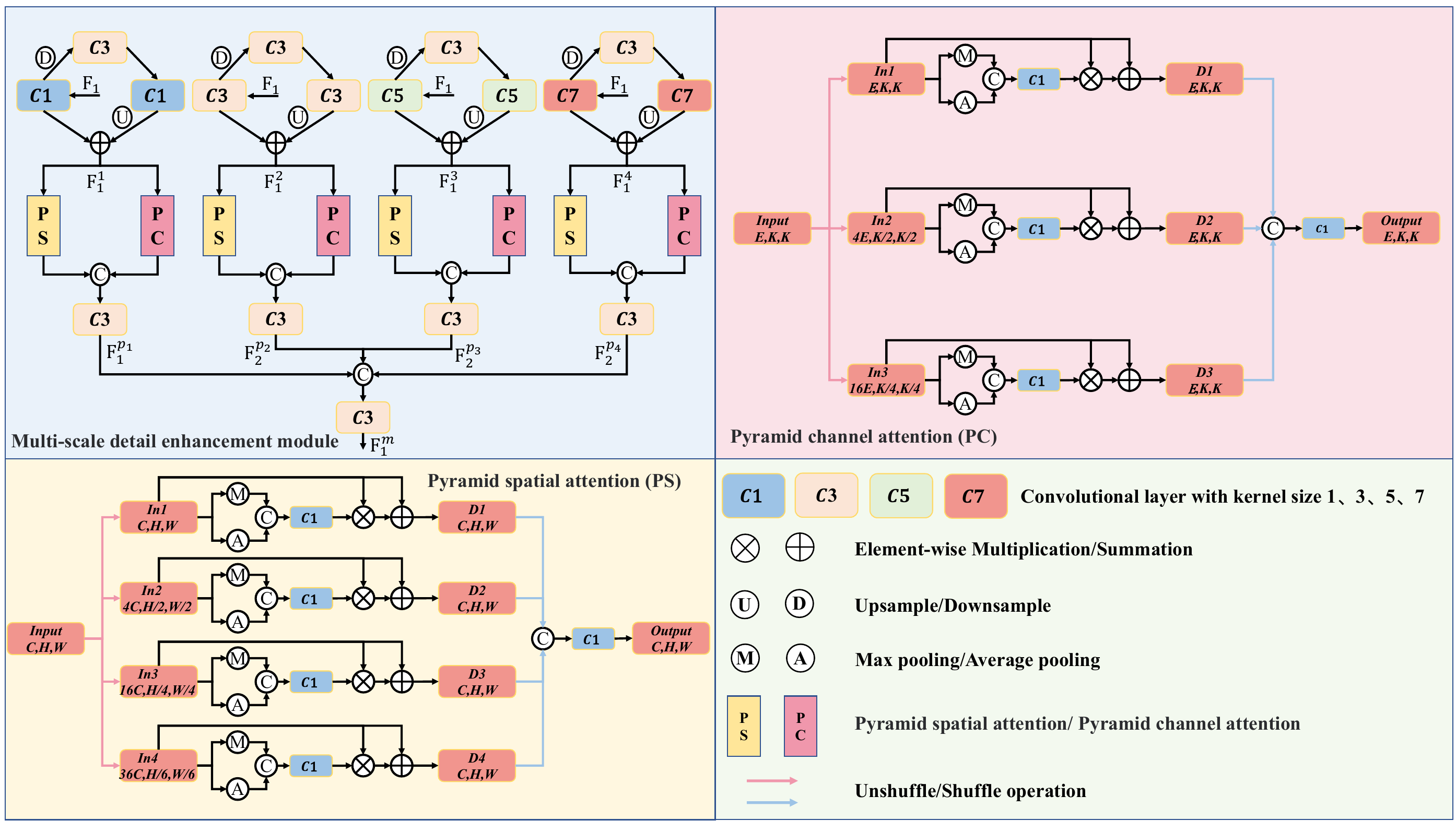}
	\caption{\small{Architecture of the multi-scale detail enhancement module.}} 
	\label{fig_mde}
\end{figure}
\subsection{Multi-scale Detail Enhancement Module}
For low-level features, most existing methods adopt ASPP-like structures to extract detailed information from single-scale features using convolutional layers with different kernel sizes. However, in optical remote sensing image (ORSI) tasks, objects often exhibit significant scale variations, directly transferring these methods to remote sensing images fails to enable the detection network to effectively learn detailed information of objects at different scales. Therefore, to address the above issue, we design the multi-scale detail enhancement (MDE) module. As shown in Fig. \ref{fig_mde}, taking the first-layer feature $\mathbf{F}_1$ as an example, the feature is first fed into four simplified U-Net branches, each employing convolutional layers with different kernel sizes to capture features with varying receptive fields. In addition, by leveraging downsampling and upsampling operations within these branches, the network is encouraged to extract fine details from low-level features at multiple scales, thus effectively enhancing the network's ability to represent objects with large scale variations,

\begin{equation}
	\mathbf{F}_1^i =Up(f_{2i-1}( f_{3}(Do(f_{2i-1}(\mathbf{F}_1)))))+f_{2i-1}(\mathbf{F}_1)
\end{equation}
where $f_{2i-1}$ denotes the convolutional layer with kernel size $2i-1$, $Do$ and $Up$ denote the downsampling and upsampling operations, respectively. Through the above approach, structural detail information can be further extracted; however, it inevitably contains irrelevant noise, requiring specific strategies for filtering. Most existing methods utilize a combination of spatial and channel attention mechanisms for feature refinement, but these techniques typically operate at a single scale and thus fail to effectively handle objects of varying scales. To end this, we design the pyramid spatial attention block and pyramid channel attention block to complete feature optimization. To be specific, as shown in Fig. \ref{fig_mde}, in the pyramid spatial attention block,  first, to obtain feature information at different scales, instead of using average pooling operations that cause information loss, we apply four unshuffle operations with different downscaling ratios (\emph{i.e.}, 1, 2, 4, 6) to the input feature $Input$. This approach not only compresses the feature scales but also preserves complete feature information, resulting in four distinct feature representations $\{\mathbf{In}_i\}_{i=1}^4$,
\begin{equation}
	\begin{cases}
		\mathbf{In}_1 = US_1(Input)\\
	\mathbf{In}_2 = US_2(Input)\\
	\mathbf{In}_3 = US_4(Input)\\
	\mathbf{In}_4 = US_6(Input)\\
	\end{cases},
\end{equation}
where $US_i$ denotes the unshuffle operation with factor$i$, where the factor specifies how many pixels along each spatial dimension are rearranged into the channel dimension. Specifically, an unshuffle with factor i reduces the spatial resolution by a factor of $i$ (\emph{i.e.,} $H \times W \rightarrow \frac{H}{i} \times \frac{W}{i}$) while increasing the channel dimension by $i^2$. The multi-scale features obtained through this process retain most of the original information, effectively preventing the loss that commonly occurs when using bilinear downsampling. Next, to refine the detailed information in the features $\{\mathbf{In}_i\}_{i=1}^4$, global average pooling and global max pooling are applied along the channel dimension. The resulting two single-channel feature maps are concatenated and processed with a 1$\times$1 convolution to generate a feature weight map which is then used to guide the detection network to focus on object details through element-wise multiplication and addition with the features $\mathbf{In}_i$. Besides, the shuffle operation is employed to restore the optimized features to their original size, yielding features $\{\mathbf{D}_i\}_{i=1}^4$ which are concatenated along the channel dimension and generate the output feature $Output$ through the 3$\times$3 convolutional layer,
\begin{equation}
	\begin{cases}
		\mathbf{D}_i = SH_{2i-1}(f_{1}(Cat(Max(\mathbf{In}_i),Avg(\mathbf{In}_i)))\times\mathbf{In}_i+\mathbf{In}_i)\\
		Output = f_{3}(Cat(	\mathbf{D}_1,	\mathbf{D}_2,	\mathbf{D}_3,	\mathbf{D}_4))
	\end{cases},
\end{equation}
where $Max$ and $Avg$ denote global max pooling and global average pooling along the channel dimension, respectively, which extract the most dominant and the average response information from the feature map. These two pooling operations generate two single-channel spatial weight maps that characterize the importance of each spatial position in terms of its maximal and average responses. $Cat$ is the concatenation operation and $SH$ is the shuffle operation, which reverses the spatial-to-channel rearrangement performed by the unshuffle operation. Specifically, shuffle redistributes the expanded channel dimensions back into the spatial dimensions, restoring the multi-scale features obtained via unshuffle to their original spatial resolution. In addition, for the pyramid channel attention block, as shown in Fig. \ref{fig_mde}, unlike the pyramid spatial attention block, which can directly perform multi-scale sampling along the spatial dimension, it is not straightforward to apply this operation on the channel dimension. To enable such sampling, we first perform a dimensional transformation on the input features $Input$. Specifically, the input feature $Input$ $\in R^{C\times H\times W}$ is first reshaped into lower-dimensional representations $\in R^{C\times HW}$ and then transposed to convert their dimensions to $ R^{HW\times C}$. Subsequently, we flatten the last dimension of the features into $K \times K$, resulting in a feature shape of $E \times K \times K$, where $E = H \times W$ and $K$ is the square root of $C$,
\begin{equation}
	Input_t=Fl((Re(Input))^T)
\end{equation}
where $Re$ denotes the reshape operation, which flattens the height and width dimensions into a single spatial dimension to obtain a lower-dimensional feature representation. The operator $()^T$ represents the transpose operation, used to swap the channel dimension with the fused spatial dimension in preparation for subsequent channel-wise processing. $Fl$ denotes the flatten operation along the channel dimension, expanding the single channel axis into a two-dimensional structure, which facilitates the following multi-scale operations applied along the channel dimension. After that, we can perform multi-scale transformations along the channel dimension.  Considering that the channel number of encoder features is compressed to 64 via a 1$\times$1 convolution, the value of K is set to 8. Therefore, when performing multi-scale transformations, we adopt unshuffle operations with scales of 1, 2, and 4 to obtain the corresponding feature representations $\{\mathbf{In}_i\}_{i=1}^3$,
\begin{equation}
	\begin{cases}
		\mathbf{In}_1 = US_1(Input_t)\\
		\mathbf{In}_2 = US_2(Input_t)\\
		\mathbf{In}_3 = US_4(Input_t)\\
	\end{cases}.
\end{equation}

Next, we apply the same refinement operations as in the pyramid spatial attention block.  After that, the three features $\{\mathbf{D}_i\}_{i=1}^3$ are fused through concatenation followed by a convolution operation. Then, we restore the features to their original size using shuffle, transpose, and dimensional transformation operations,
\begin{equation}
	\begin{cases}
		\mathbf{D}_i=SH_{2i-1}(f_1(Cat(Max(\mathbf{In}_i),Avg(\mathbf{In}_i)))\times\mathbf{In}_i+\mathbf{In}_i)
		\\
		Output = Fl((Re(f_{3}(Cat(	\mathbf{D}_1,	\mathbf{D}_2,	\mathbf{D}_3))))^T)
	\end{cases},
\end{equation}
where $SH_{2i-1}$ denotes shuffle operation, which restores the feature resolution from [$(2i-1)^2E,K/(2i-1),K/(2i-1)$] to [$E,K,K$], $Re$ reshapes the feature dimension from [$E,K,K$] to low dimension [$E,K\times K$]($K\times K=C$) and $()^T$ performs the transpose operation on the feature dimensions. Then, $Fl$  flattens the spatial dimensions of the feature map to obtain a feature representation of size [$C,H,W$]. Finally, the output features from the pyramid spatial attention and pyramid channel attention in the four branches are first fused through concatenation followed by a 3$\times$3 convolution. The resulting four features $\{\mathbf{F}_{1}^{p_j}\}_{j=1}^4$ are then further concatenated and fused with another 3$\times$3 convolution to produce the final output feature $\mathbf{F}_1^m$ of the MDE module,
\begin{equation}
	\mathbf{F}_1^m = f_{3}(Cat(\mathbf{F}_1^{p_1},\mathbf{F}_1^{p_2},\mathbf{F}_1^{p_3},\mathbf{F}_1^{p_4})).
\end{equation}
\subsection{Dual-branch Geo-Gran Complementary Module}
\begin{figure}[!t]
	\centering
	\includegraphics[width=0.49\textwidth]{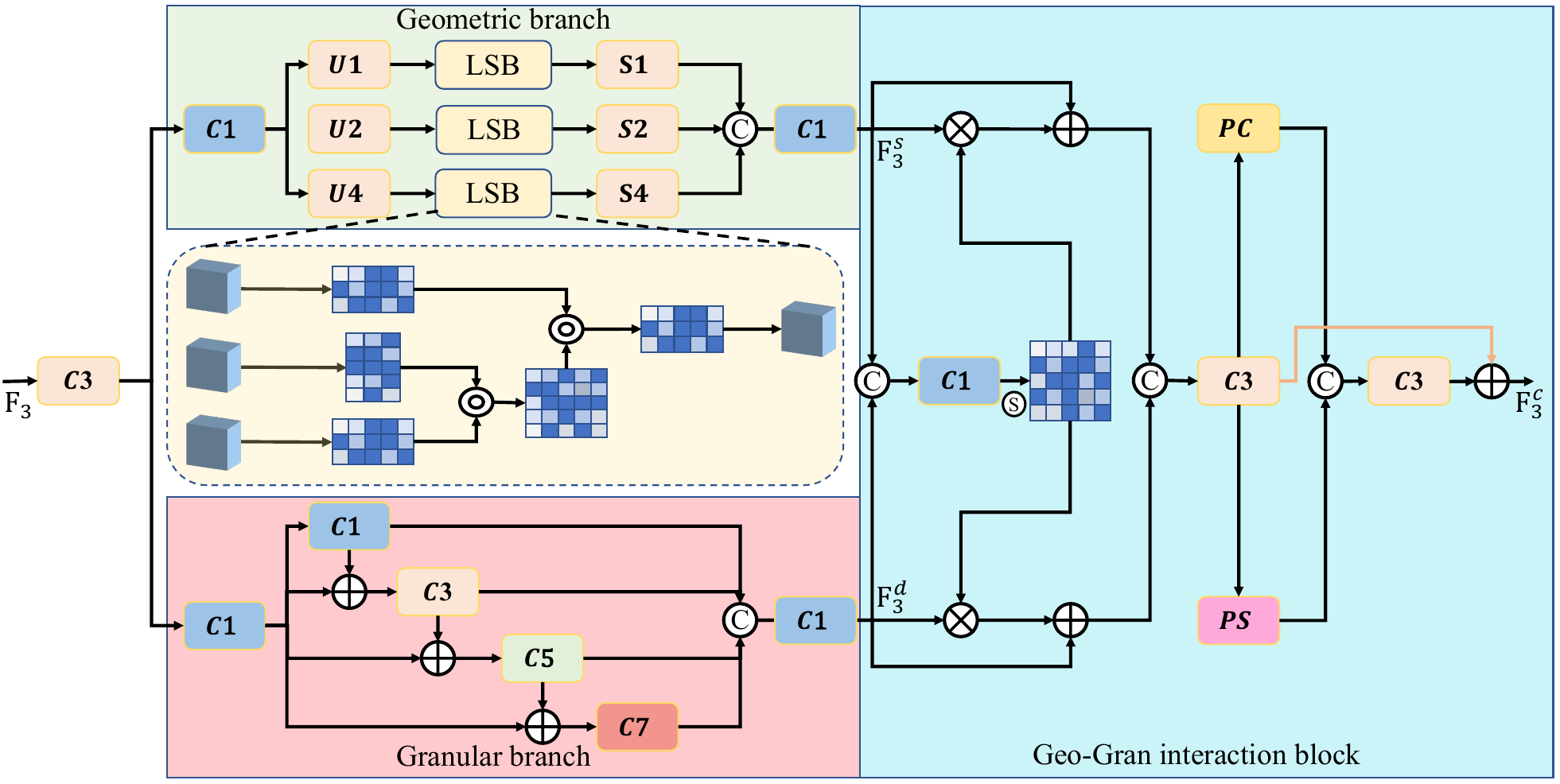}
	\caption{\small{Architecture of the dual-branch geo-gran complementary module.}} 
	\label{fig_dgc}
\end{figure}

As described in the introduction, most existing methods utilize cross-attention mechanisms to explore correlations among mid-level features, yet they overlook the fact that mid-level features also contain valuable detail and positional information, which can provide strong support for remote sensing object detection. To fully exploit those information, as shown in Fig. \ref{fig_dgc}, we design the dual-branch geo-gran complementary (DGC) module, which consists of geometric branch, granular branch, and a  geo-gran interaction block for fusing the features from both branches. To be specific, taking feature $\mathbf{F}_3$ as an example, we first expand the input feature dimension from 64 to 128 using a 3$\times$3 convolution, where one half is used as the input to granular branch, and the other half is assigned to geometric branch for location information extraction. For the granular branch, four sub-branches with different convolutional kernel sizes are employed to extract fine details under varying receptive fields. Unlike other methods that directly fuse features from each branch, we iteratively feed features from smaller receptive fields into the subsequent branch as input, enabling the network to progressively enhance detail extraction. Finally, the features from all four sub-branches are fused to obtain the granular feature $\mathbf{F}_3^d$,
\begin{equation}
	\begin{cases}
		\mathbf{In}_1 = f_{1}(Input)\\
		\mathbf{In}_2 = f_{3}(Input+\mathbf{In}_1)\\
		\mathbf{In}_3 = f_{5}(Input+\mathbf{In}_2)\\
		\mathbf{In}_4 = f_{7}(Input+\mathbf{In}_3)\\
		\mathbf{F}_3^d = f_{1}(Cat(\mathbf{In}_1,\mathbf{In}_2,\mathbf{In}_3,\mathbf{In}_4))
	\end{cases}.
\end{equation}

For the geometric branch, considering that low-scale features often contain relatively accurate positional information, most existing methods employ average pooling to downsample features for acquiring such information. However, this approach inevitably leads to information loss. To address this, in the geometric branch, we adopt three different levels of pixel unshuffle operations on the input feature $Input$,which reorganize spatial information into the channel dimension and therefore reduce the spatial resolution without losing feature content, obtaining three features $\{\mathbf{In}_i\}_{i=1}^3$,
\begin{equation}
	\begin{cases}
		\mathbf{In}_1 = US_1(Input)\\
		\mathbf{In}_2 = US_2(Input)\\
		\mathbf{In}_3 = US_4(Input)\\
	\end{cases}.
\end{equation}

Subsequently, we apply a location sensing  block to each feature to optimize the positional information through a self-attention mechanism. As illustrated by the yellow regions in Fig. \ref{fig_dgc}, feature  $\mathbf{In}_i$ is projected into three separate feature representations $\mathbf{In}_i^q$, $\mathbf{In}_i^k$ and $\mathbf{In}_i^v$ by the $1\times1$ convolutional layers. Then, we reshape those features into low-dimension feature, where $\mathbf{\hat{In}}_i^q$ and $\mathbf{\hat{In}}_i^v$ $\in R^{HW\times C}$, while $\mathbf{\hat{In}}_i^k$ $\in R^{C\times HW}$. Next, $\mathbf{\hat{In}}_i^k$ and $\mathbf{\hat{In}}_i^q$ and combined by the matrix multiplication to generate response map $\mathbf{M}_i$ $\in R^{C\times C}$ which is imposed on $\mathbf{\hat{In}}_i^v$ by the  matrix multiplication followed by the dimension transformation to obtain location-aware feature,
\begin{equation}
	\begin{cases}
		\mathbf{In}_i^q,\mathbf{In}_i^k, \mathbf{In}_i^v = f_{1}(\mathbf{In_i})\\
		
		\mathbf{M}_i=\mathbf{\hat{In}}_i^k\otimes \mathbf{\hat{In}}_i^q\\
		
		\mathbf{In}_i^{s} = DT(\mathbf{\hat{In}}_i^v\otimes \mathbf{M}_i)
	\end{cases}.
\end{equation}
where $\otimes$ denotes the matrix multiplication and $DT$ is the dimensional transformation, which first transposes the feature from $HW \times C$ to $C \times HW$, and then restores it to its original shape $C \times H \times W$ via a reshape operation. Subsequently, we restore the scales of the three optimized features $\{\mathbf{In}_i\}_{i=1}^3$ using shuffle operations, and then fuse them through concatenation and convolution to obtain the final positional feature $\mathbf{F}_3^s$,
\begin{equation}
	\mathbf{F}_3^s = f_{3}(Cat(SH_1(\mathbf{In}_1^{s}),SH_2(\mathbf{In}_2^{s}),SH_4(\mathbf{In}_3^{s)})).
\end{equation}

To fully fuse the geometric feature $ \mathbf{F}_3^s$ and granular feature $\mathbf{F}_3^d$, we design a geo-gran interaction block that leverages their shared responses to achieve mutual enhancement. The geometric branch provides stable global structural cues, whereas the granular branch contributes fine-grained textures and boundary details. By computing a shared activation, the model identifies spatial locations where both branches consistently respond, enabling complementary information from the two subspaces to be jointly reinforced. Specifically, as shown in Fig. \ref{fig_dgc}, the two features are concatenated and processed by a $1\times1$ convolution followed by a sigmoid function to generate a single-channel weight map $\mathbf{W}_3$. This weight map explicitly encodes the complementary strength between geometric and granular cues, highlighting regions where location and detail information are mutually supportive. It is then propagated back to both branches through element-wise multiplication and addition, producing the enhanced features $\mathbf{F}_3^{se}$ and $\mathbf{F}_3^{de}$,
\begin{equation}
	\begin{cases}
		\mathbf{W}_3 = \sigma(f_{1}(Cat(\mathbf{F}_3^s,\mathbf{F}_3^d)))\\
		\mathbf{F}_3^{se} = \mathbf{F}_3^s\times	\mathbf{W}_3 +  \mathbf{F}_3^s\\
		\mathbf{F}_3^{de} = \mathbf{F}_3^d\times	\mathbf{W}_3 +  \mathbf{F}_3^d
	\end{cases},
\end{equation}
where $\sigma$ denotes the sigmoid activation function. Finally, the two enhanced features are fused again and then separately processed by the pyramid spatial attention and pyramid channel attention blocks. The resulting two features are concatenated and fused using a 3$\times$3 convolution, and finally added to feature $\mathbf{F}_3^{sde}$ to obtain the output feature $\mathbf{F}_3^c$ of DGC module,
\begin{equation}
	\begin{cases}
	\mathbf{F}_3^{sde }= f_{3}(Cat(	\mathbf{F}_3^{se},	\mathbf{F}_3^{de}))\\
	\mathbf{F}_3^c = \mathbf{F}_3^{sde} + f_{3}(Cat(PC(\mathbf{F}_3^{sde}),PS(\mathbf{F}_3^{sde})))
	\end{cases},
\end{equation}
where $PS$ and $PC$ denote pyramid spatial attention and pyramid channel attention, respectively.

\subsection{Deep Semantic Perception  Module }
\begin{figure}[!t]
	\centering
	\includegraphics[width=0.4\textwidth]{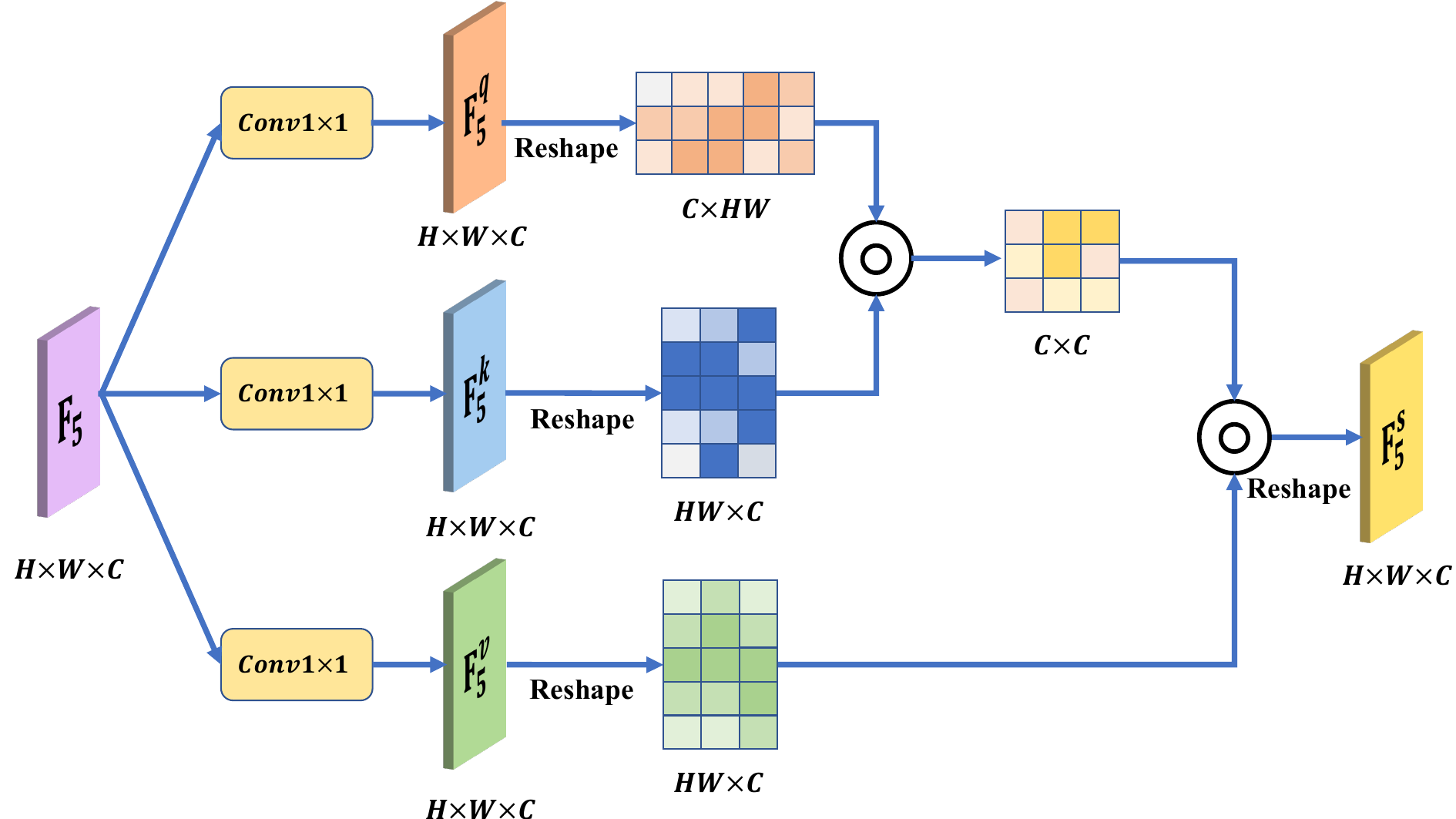}
	\caption{\small{Architecture of the deep semantic perception module.}} 
	\label{fig_dsp}
\end{figure}

Considering that the fifth-layer feature has a relatively small scale of 64$\times$12$\times$12, we do not apply scale reduction operations to this feature but instead use the same self-attention operation as in semantic perception block. Therefore, we provide only a brief introduction to the DSP module. As shown in Fig. \ref{fig_dsp}, feature  $\mathbf{F}_5$ is processed with the $1\times1$ convolutional layers to obtain $\mathbf{F}_5^q$, $\mathbf{F}_5^k$ and $\mathbf{F}_5^v$. Then, the low-dimensional features $\mathbf{\hat{F}}_5^k$ and $\mathbf{\hat{F}}_5^q$ are combined via matrix multiplication to generate the response map $\mathbf{M}_5 \in \mathbb{R}^{C \times C}$, which is subsequently applied to $\mathbf{\hat{F}}_5^v$ through another matrix multiplication followed by a dimension transformation to obtain the location feature $\mathbf{F}_5^s$,
\begin{equation}
	\begin{cases}
		\mathbf{F}_5^q,\mathbf{F}_5^k, \mathbf{F}_5^v = f_{1}(\mathbf{F_5})\\
		
		\mathbf{M}_5=\mathbf{\hat{F}}_5^k\otimes \mathbf{\hat{F}}_5^q\\
		
		\mathbf{F}_5^{s} = DT(\mathbf{\hat{F}}_5^v\otimes \mathbf{M}_5)
	\end{cases}.
\end{equation}

\subsection{Local-Global Guidance Fusion Module }
\begin{figure}[!t]
	\centering
	\includegraphics[width=0.4\textwidth]{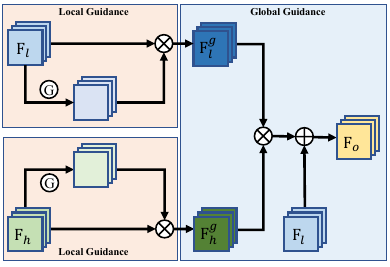}
	\caption{\small{Architecture of the local-global guidance fusion module.}} 
	\label{fig_lgd}
\end{figure}

In the field of object detection, the mainstream detection framework is the encoder-decoder structure, primarily based on U-Net. However, most existing methods use only a single convolutional layer in the decoding process to transmit multi-level feature information from bottom to top. This approach fails to sufficiently guide the network to focus on target information; therefore, we design the local-global guidance fusion (LGF) module to address this limitation and achieve full integration of detailed and positional information. As shown in the Fig. \ref{fig_lgd}, the LGF module consists of two stages: local guidance and global guidance. In the local guidance stage, a gate convolution is applied to each pair of adjacent features (\emph{i.e.,} $\mathbf{F}_l$ and $\mathbf{F}_h$) to obtain local guidance features, which are then multiplied with their corresponding input features to achieve local guidance,
\begin{equation}
	\begin{cases}
		\mathbf{F}_l^g = f_{3}(\mathbf{F}^l)+\mathbf{F}^l\\
			\mathbf{F}_h^g = f_{3}(\mathbf{F}^h)+\mathbf{F}^h\\
	\end{cases}.
\end{equation}

Subsequently, to enable the high-level feature $	\mathbf{F}_h^g$ to guide the low-level feature 	$\mathbf{F}_l^g$, we perform element-wise multiplication between the two features while incorporating the original high-level feature $\mathbf{F}_h$ to preserve some initial feature information during this process,
\begin{equation}
\mathbf{F}_o = 	\mathbf{F}_h^g \times 	\mathbf{F}_l^g +\mathbf{F}_h.
\end{equation}

\subsection{Deep Supervision}
To enhance the performance of G$^2$HFNet, we adopt a fusion loss comprising binary cross-entropy (BCE) loss \cite{de2005tutorial}, boundary intersection-over-union (IOU) loss \cite{rahman2016optimizing}, and F-measure (FM) loss \cite{zhao2019optimizing} to jointly supervise the five saliency predictions. The total loss function $L_{total}$ is defined as follows:
\begin{equation}
	L_{total}=\frac{1}{N} \sum_{i=1}^{N}(L_{bce}+L_{iou}+L_{fm}),
\end{equation}
where $N$ is batch size of training phase.
\subsubsection{BCE Loss}
\begin{equation}
	L_{bce}=-\sum_{(x,y)}[\mathbf{G}(x,y)log(\mathbf{S}(x,y))+\bar{\mathbf{G}}(x,y)log(\bar{\mathbf{S}}(x,y))] ,
\end{equation}
where $\mathbf{S}$ and $\mathbf{G}$ are the predicted result and saliency groundtruth. $\bar{\mathbf{G}}=1-\mathbf{G}$ and  $\bar{\mathbf{S}}=1-\mathbf{S}$.
\subsubsection{IOU Loss}
\begin{equation}
	L_{iou}= 1-\frac{\sum_{(x,y)}\mathbf{S}(x,y)\mathbf{G}(x,y)}{\sum_{(x,y)}[\mathbf{S}(x,y)+\mathbf{G}(x,y)-\mathbf{S}(x,y)\mathbf{G}(x,y)]}.
\end{equation}
\subsubsection{FM Loss}
\begin{equation}
	L_{fm}=1-\frac{(1+\beta^2)TP_{i}}{H_i},
\end{equation}
where $H=\beta^2(TP+FN)+(TP+FP)$, $TP$, $FP$, and $FN$ denote true positive, false positive, and false negative respectively.


\begin{figure*}
	\centering
	\includegraphics[width=0.83\linewidth]{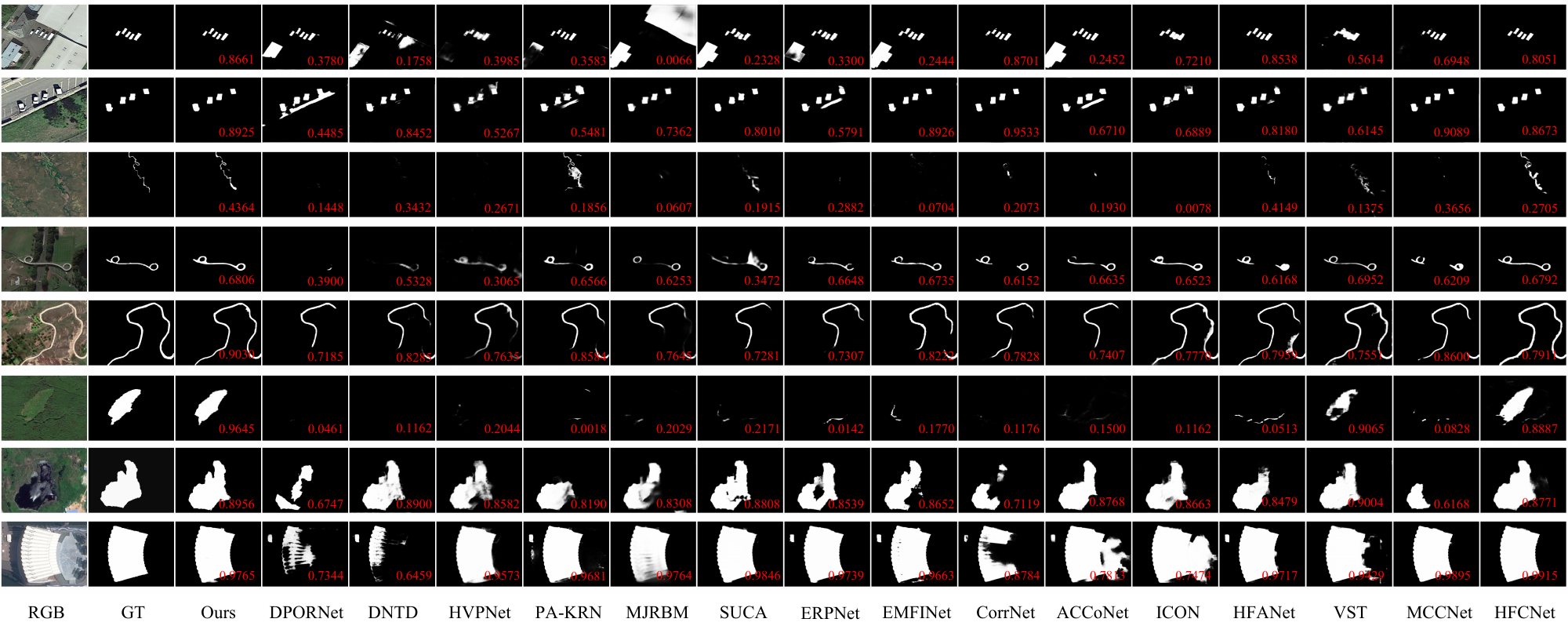}
	\caption{\small{Visual comparison of EORSSD, ORSSD and ORSI-4199 datasets.}} 
	\label{fig_eor}
\end{figure*}

\begin{table*}[t]
\small
\renewcommand{\arraystretch}{0.75}
\setlength\tabcolsep{10pt}
\centering
\caption{Quantitative comparisons with 18 methods on EORSSD, ORSSD and ORSI-4199.}
\begin{tabular}{c|ccc|ccc|ccc}
	\hline\hline
	Dataset & \multicolumn{3}{c|}{EORSSD} &\multicolumn{3}{c}{ORSSD}&\multicolumn{3}{c}{ORSI-4199}\\
	\hline
	Metric& $M$$\downarrow$ & $F_{\beta}$$\uparrow$ & $E_{\xi}$  $\uparrow$
	& $M$$\downarrow$ & $F_{\beta}$$\uparrow$ & $E_{\xi}$  $\uparrow$& $M$$\downarrow$ & $F_{\beta}$$\uparrow$ & $E_{\xi}$  $\uparrow$ \\
	\hline
	PoolNet   & 0.0210 & 0.4614 & 0.6853 & 0.0359 & 0.6141 & 0.8142& 0.0543 & 0.7353  & 0.8550   \\
	
	CSNet     & 0.0171 & 0.6284  & 0.8325  & 0.0188 & 0.7574  & 0.9049 & 0.0525 & 0.7151  & 0.8442    \\
	
	R3Net     & 0.0171 & 0.4174  & 0.6472 & 0.0401 & 0.7320  & 0.8713  & 0.0403 & 0.7769 & 0.8828 \\
	
	DPORTNet  & 0.0151 & 0.7377 & 0.8788  & 0.0222 & 0.7861  & 0.8940& 0.0571 & 0.7514  & 0.8641  \\
	
	
	DNTD      & 0.0114 & 0.7192  & 0.8714 & 0.0219 & 0.7598  & 0.8862 & 0.0426 & 0.8032  & 0.9019    \\
	
	HVPNet    & 0.0111 & 0.6172  & 0.8098 & 0.0227 & 0.6701  & 0.8485 & 0.0420 & 0.7638  & 0.8845  \\
	
	
	
	PA-KRN    & 0.0106 & 0.7773  & 0.9186  & 0.0141 & 0.8430 & 0.9373& 0.0384 & 0.8161  & 0.9157  \\
	MJRBM     & 0.0101 & 0.6994  & 0.8861  & 0.0166 & 0.7951  & 0.9308  & 0.0376 & 0.7975  & 0.9079 \\
	SUCA      & 0.0098 & 0.7161  & 0.8747  & 0.0147 & 0.7693  & 0.9068& 0.0306 & 0.8375  & 0.9256  \\
	ERPNet    & 0.0091 & 0.7405  & 0.9153 & 0.0138 & 0.8245  & 0.9489 & 0.0359 & 0.7991  & 0.9025  \\
	EMFINet   & 0.0086 & 0.7752  & 0.9393  & 0.0112 & 0.8461  & 0.9616& 0.0332 & 0.8136 & 0.9123   \\
	CorrNet   & 0.0085 & \textcolor{blue}{0.8092}  & \textcolor{blue}{0.9533}  & 0.0101 & \textcolor{green}{0.8720}  & \textcolor{green}{0.9722} & 0.0368 & 0.8489  & 0.9296 \\
	ACCoNet   & 0.0076 & 0.7809  & 0.9408   & 0.0092 & 0.8651 & 0.9716& 0.0316 &\textcolor{green}{0.8539} & 0.9393  \\
	ICON      & 0.0074 & 0.7861  & 0.9139  & 0.0119 & 0.8327  & 0.9386  & 0.0284 & 0.8497  & \textcolor{blue}{0.9432}  \\
	HFANet    & 0.0073 & \textcolor{green}{0.8082}  &0.9243& 0.0096& 0.8668 & 0.9486 & 0.0316 & 0.8271  & 0.9184  \\
	VST       & 0.0069 & 0.7010  & 0.8744 & 0.0096 & 0.8189 & 0.9370& \textcolor{green}{0.0283} & 0.7929  & 0.9059  \\
	MCCNet    &  \textcolor{green}{0.0068} & 0.7936 & \textcolor{green}{0.9470} & \textcolor{green}{0.0091} & \textcolor{blue}{0.8808}  & \textcolor{blue}{0.9741}& 0.0318 & \textcolor{blue}{0.8550} & \textcolor{green}{0.9393}  \\
	HFCNet    & \textcolor{blue}{0.0051} & 0.7845 & 0.9280& \textcolor{blue}{0.0073} & 0.8581 & 0.9554  &\textcolor{blue}{0.0270}&0.8272&0.9234\\
	Ours      & \textbf{\textcolor{red}{0.0041} }& \textbf{\textcolor{red}{0.8808}}  & \textbf{\textcolor{red}{0.9807}} & \textbf{\textcolor{red}{0.0056} }& \textbf{\textcolor{red}{0.9147}} & \textbf{\textcolor{red}{0.9868}}&\textbf{\textcolor{red}{0.0242} }& \textbf{\textcolor{red}{0.8862}}& \textbf{\textcolor{red}{0.9557} }   \\
	\hline
\end{tabular}
\label{tab_comparison}
\end{table*}

\section{Experiments And Analyses}
\subsection{Implementation Details and Evaluation Metrics}
\subsubsection{Implementation Details}
We evaluate our G$^2$HFNet on three public datasets: ORSSD \cite{li2019nested}, EORSSD \cite{zhang2020dense}, and ORSI-4199 \cite{tu2021orsi}. ORSSD, the first publicly available dataset for ORSI-SOD, contains 800 images with pixel-level ground truths (GTs), partitioned into 600 training images and 200 testing images. EORSSD includes 2,000 images with GTs, divided into 1,400 for training and 600 for testing. ORSI-4199, the largest ORSI-SOD dataset, consists of 4,199 images with GTs, with 2,000 used for training and 2,199 for testing.

We implement G$^2$HFNet using the PyTorch framework and conduct all experiments on a workstation equipped with a single NVIDIA RTX 4080 GPU. We adopt the Swin Transformer as the backbone network, initializing it with pre-trained parameters. During training, all images are resized to $384 \times 384$, and we apply data augmentation techniques, including random flipping, cropping, and rotation. The network is optimized using the RMSprop optimizer \cite{tieleman2012rmsprop}, with an initial learning rate of $1 \times 10^{-4}$, a momentum of 0.9, a batch size of 4 and the learning rate decays to 0.7 every 12 epochs after 30 epochs, and the training epoch is 80.

\subsubsection{Evaluation Metrics}
To assess the performance of the proposed G$^2$HFNet, we adopt five commonly used evaluation metrics: mean absolute error ($M$) \cite{perazzi2012saliency}, F-measure ($F_{\beta}$) \cite{achanta2009frequency}, and E-measure ($E_{\xi}$) \cite{fan2018enhanced}. 
%




\subsection{Comparison with State-of-the-Art Methods}
In experiments, we compare the proposed G$^2$HFNet with other state-of-the-art NSI-SOD and ORSI-SOD methods, including R3Net \cite{deng2018r3net}, PoolNet \cite{liu2019simple}, CSNet \cite{gao2020highly}, SUCA \cite{li2020stacked}, PA-KRN \cite{xu2021locate}, VST \cite{liu2021visual}, DPORTNet \cite{liu2022disentangled}, DNTD \cite{fang2022densely}, ICON \cite{zhuge2022salient}, MJRBM \cite{tu2021orsi}, EMFINet \cite{wang2022multiscale}, ERPNet \cite{zhou2022edge}, ACCoNet \cite{li2022adjacent}, CorrNet \cite{li2023lightweight}, MCCNet \cite{li2021multi},  HFANet \cite{wang2022hybrid}, and HFCNet \cite{liu2024heterogeneous}.  To ensure fairness in comparison, we utilized the source code provided by the original authors.

\subsubsection{Qualitative Comparison}
To demonstrate the capability of our G$^2$HFNet, we present visualizations from challenging scenes (\emph{i.e.}, big salient object (BSO), narrow salient object (NAO), multiple small salient object (MSSO), low contrast object (LCO)) from EORSSD \cite{zhang2020dense}, ORSSD \cite{li2019nested} and ORSI-4199 \cite{tu2021orsi} in Fig. \ref{fig_eor}, where the $F_{\beta}$ score of each example is shown at the bottom-right corner of its corresponding image.
\paragraph{Advantages in Big Salient Object}
In ORSI salient object detection, scenes containing large objects often present significant challenges. For instance, in the 7$^{th}$ row of  Fig. \ref{fig_eor}, a massive lake occupies the majority of the image, and its intricate contours pose substantial difficulties for detection. From the  visualization results, it is clearly observed from the Fig. \ref{fig_eor} that none of the other methods can accurately highlight the complete contours of the lake. In particular, methods DPORTNet(0.6747), PA-KRN(0.8190) and MCCNet(0.6168)  are only able to identify partial regions within the lake, failing to capture its overall structure. Besides, in the 8$^{th}$ row of  Fig. \ref{fig_eor}, the stadium roof occupies more than one-third of the image area. Similarly, several methods (\emph{e.g.}, DPORTNet(0.7344) and DNTN(0.8900)) show missing regions in their detection results for this area, while most methods (\emph{e.g.},  PA-KRN(0.7344), EPRNet(0.8539), EMFINet(0.8652) and HFANet(0.8479)) mistakenly identify the small buildings next to the stadium as salient regions. This limitation is attributable to the inherent challenges these methods encounter in accurately capturing shallow, fine-grained detail features.


\paragraph{Advantages in Narrow Salient Object}

Due to the relatively high imaging altitudes of remote sensing platforms, objects often appear extremely narrow and spatially compressed within the captured images. This pronounced reduction   considerably increases the difficulty of distinguishing these objects from complex backgrounds, thereby posing substantial challenges for accurate detection and precise segmentation. As shown in Fig. \ref{fig_eor} ($3^{rd}$, $4^{th}$ and $5^{th}$ rows), we provide three visualizations from datasets. For example, in the $3^{rd}$ row of Fig. \ref{fig_eor}, there is an extremely narrow and winding river. Most comparison methods (\emph{e.g.}, DPORTNet, DNTN, HVPNet, EPRNet and ICON) fail to detect any part of the river region, while PA-KRN, SUCA, VST and HFCNet are able to identify certain segments but produces very blurred results. Additionally, in the $5^{th}$ row, there is a narrow road that crosses the entire image. As observed from the detection results, most methods exhibit missing segments in the visualization of this road. Additionally, the corresponding $F_{\beta}$ scores further demonstrate that some methods fail to effectively recognize narrow targets (\emph{e.g.}, DPORNet(0.7158), MJRBM(0.7645), ICON(0.7779) and Our(0.9030)). This phenomenon arises from the fact that some methods lack multi-receptive-field information, hindering their capability to integrate long-range object information effectively.

\paragraph{Advantages in Multiple Small Salient Object}
In remote sensing imagery, the high acquisition altitude often causes multiple targets to appear very small within the image, posing substantial difficulties for accurate detection.  In Fig. \ref{fig_eor} ($1^{st}$ and $2^{nd}$ rows), multiple cars are parked together. Due to their small sizes, many methods (\emph{e.g.},  DPORTNet, MJRBM and ACCoNet) incorrectly detect nearby buildings as salient objects. While certain methods (\emph{e.g.,} HVPNet and EPRNet) successfully identify the main structures of the cars, the resulting detections remain notably blurred. In addition, the $F_{\beta}$ scores annotated in the comparison figures also demonstrate that our method (0.8925) achieves superior detection performance compared with other approaches (\emph{e.g.}, DPORNet(0.4485), HVPNet(0.5267) and VST(0.6145)) when handling multiple small salient objects.


\paragraph{Advantages in Low Contrast Object}
In the field of object detection, low-contrast targets have consistently posed a significant challenge. As illustrated in the $6^{th}$ row of Fig. \ref{fig_eor}, the central region exhibits a subtle contrast against the surrounding dense vegetation, presenting a typical low-contrast scenario in optical remote sensing imagery. In this scenario, we observe that almost all comparison methods fail to identify the target region in the image, with only VST(0.9065) and HFCNet(0.8887) managing to detect partial areas of the object, whereas our method(0.9654) achieves clearly superior results even under such low-contrast conditions.


As evidenced by the visualization results, our method demonstrates exceptional robustness and consistently outperforms other approaches in various challenging scenarios.

\begin{figure*}[!t]
	\centering
	\includegraphics[width=0.83\textwidth]{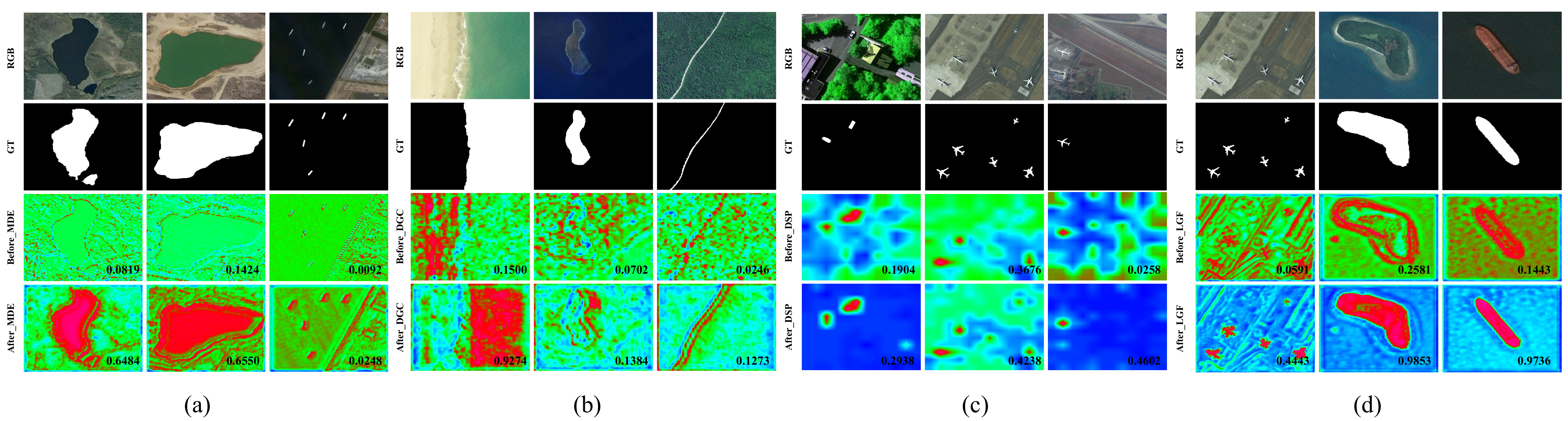}
	\caption{\small{Illustration of the impact of MDE module (a), DGC module (b), DSP module (c) and LGF module (d).}} 
	\label{fig_hot}
\end{figure*}

\begin{figure}[!t]
	\centering
	\includegraphics[width=0.42\textwidth]{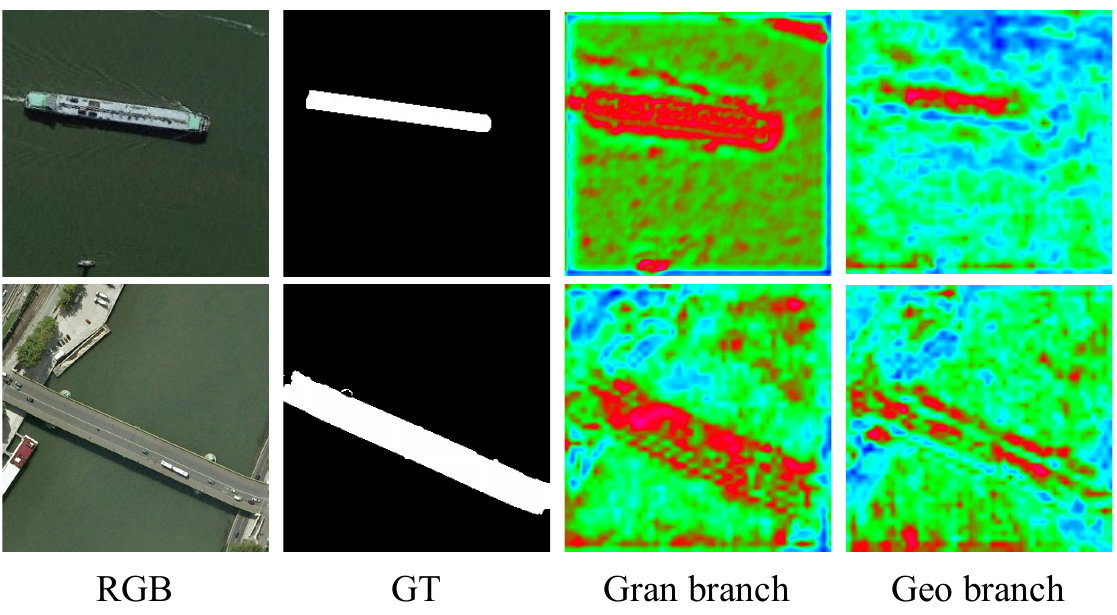}
	\caption{\small{Illustration of geometric and granular branch.}} 
	\label{fig_gg}
\end{figure}

\begin{table}[t]
	\centering
	\renewcommand{\arraystretch}{0.95}
	\setlength\tabcolsep{1.3pt}
	\caption{Comparison of the model complexity and the average running speed.}
	\begin{tabular}{c c c c c c c c c}
		\hline
		\hline
		Method&PoolNet&R3Net&PA-KRN&MJRBM&SUCA&ERPNet&EMFINet\\
		Speed(FPS)&80.9&2&16&15.5&24&73.7&25\\
		Para(M)&53.6&56.2&141.1&43.5&117.71&250.0&107.3\\
		FLOPs(G)&97.6&47.5&617.7&101.3&56.4&227.4&480.9\\
		\hline
		Method&CorrNet&ACCoNet&HFANet&VST&MCCNet&HFCNet&Ours\\
		Speed(FPS)&100&59.9&26&23&39.5&38&18.3\\
		Para(M)&4.06&273.5&60.5&83.8&67.7&140.8&95.1\\
		FLOPs(G)&21.1&368.8&68.3&23.2&234.2&120.41&94.1\\
		\hline
	\end{tabular}
	\label{tab_time}
\end{table}

\begin{table}[t]
	\small
	\renewcommand{\arraystretch}{0.9}
	\setlength\tabcolsep{8pt}
	\centering
	\caption{FLOPs and parameter of each module.}
	\begin{tabular}{ccc}
		\hline
		\multicolumn{1}{c}{Settings}&  \multicolumn{1}{c}{FLOPs(G)}&\multicolumn{1}{c}{Para(M)}\\
		\hline
		
		MDE&20.5&2.3\\
		DGC&16.3&1.4\\
		DSP&5.3&0.012\\

		\hline
	\end{tabular}
	\label{tab_module_flop}
\end{table}

\subsubsection{Quantitative comparison}  

In this section, we present a quantitative performance comparison of our G$^2$HFNet with other SOD methods using three evaluation metrics, as detailed in Table \ref{tab_comparison}. The results demonstrate that our proposed method outperforms the others across all three datasets. 

Table. \ref{tab_comparison} summarizes the quantitative comparisons between our method and 18 state-of-the-art approaches on the EORSSD, ORSSD, and ORSI-4199 datasets. The results demonstrate that our G$^2$HFNet consistently outperforms all compared methods across these benchmarks. Specifically, on the EORSSD dataset, for the $M$ metric (where lower is better), our method achieves the lowest score of 0.0041, representing a $19.6\%$ decrease over the best previous method HFCNet. In addition, G$^2$HFNet attains the best results in other metrics, achieving average improvements of $12.1\%$ in $F_{\beta}$ and $5.6\%$ in $E_{\xi}$. Compared with MCCNet, our method greatly outperforms it by $11.0\%$ on $F_{\beta}$ and $3.4\%$ on $E_{\xi}$.  Besides, on the ORSSD dataset,  our GHFNet obtains the top performance across three metrics and notably surpasses the novel method, MCCNet, by $6.6\%$ and $3.4\%$ in terms of $F_{\beta}$ and $E_{\xi}$, respectively. Additionally, on the ORSI-4199 dataset, G$^2$HFNet outperforms VST by substantial margins, with decrease of $14.4\%$ in $M$ and  improvements of $10.7\%$ in $F_{\beta}$, and $5.5\%$ in $E_{\xi}$. Furthermore, we compare the average running speed, FLOPs, and parameter counts of several representative detection methods.  As shown in Table \ref{tab_time}, our method maintains a medium-level model complexity, with FLOPs and parameters neither the highest nor the lowest among all compared approaches. This indicates that G$^2$HFNet achieves a balanced trade-off between efficiency and representational capability. Despite involving extensive matrix operations, our model still reaches an average running speed of 18 FPS. Moreover, we also provide the FLOPs and parameters of each module in Table \ref{tab_module_flop}. From the Table \ref{tab_module_flop}, it can be seen that the three modules have clearly different levels of complexity. MDE is the most computationally expensive (20.5G FLOPs, 2.3M params), DGC is moderate (16.3G, 1.4M), while DSP is extremely lightweight (5.3G, 0.012M), as it employs only three 1 $\times$ 1 convolutions, leading to a minimal number of learnable parameters. This shows that each module contributes at a different complexity level without redundancy, ensuring the overall framework remains efficient and practical.
\begin{table}[t]
\small
	\renewcommand{\arraystretch}{0.9}
	\setlength\tabcolsep{8pt}
	\centering
	\caption{Ablation study on different modules.}
	\begin{tabular}{cccc}
		\hline
		\multicolumn{1}{c}{Settings}&  \multicolumn{1}{c}{$M$$\downarrow$}&\multicolumn{1}{c}{$F_{\beta}$$\uparrow$}&  \multicolumn{1}{c}{$E_{\xi}$ $\uparrow$}\\
		\hline
		w/o MDE&0.0059&0.8724&0.9708\\
		w/o DGC&0.0054&0.8734&0.9782\\
		w/o DSP&0.0047&0.8682&0.9763\\
		w/o LGF&0.0051&0.8554&0.9701\\
		\textbf{Ours}&\textbf{0.0041}&\textbf{0.8808}&\textbf{0.9807}\\
		\hline
		
		\hline
	\end{tabular}
	\label{tab_wo}
\end{table}

\begin{table}[t]
\small
	\renewcommand{\arraystretch}{0.9}
	\setlength\tabcolsep{8pt}
	\centering
	\caption{Ablation study on MDE module.}
	\begin{tabular}{cccc}
		\hline
		\multicolumn{1}{c}{Settings}&  \multicolumn{1}{c}{$M$$\downarrow$}&\multicolumn{1}{c}{$F_{\beta}$$\uparrow$}&  \multicolumn{1}{c}{$E_{\xi}$ $\uparrow$}\\
		\hline
		w/o U&0.0057&0.8669&0.9746\\
		w/o PA&0.0058&0.8659&0.9724\\
		\textbf{Ours}&\textbf{0.0041}&\textbf{0.8808}&\textbf{0.9807}\\
		\hline
		
		\hline
	\end{tabular}
	\label{tab_mde}
\end{table}

\begin{table}[t]
	\small
	\renewcommand{\arraystretch}{0.9}
	\setlength\tabcolsep{8pt}
	\centering
	\caption{Ablation study on DGC module.}
	\begin{tabular}{cccc}
		\hline
		\multicolumn{1}{c}{Settings}&  \multicolumn{1}{c}{$M$$\downarrow$}&\multicolumn{1}{c}{$F_{\beta}$$\uparrow$}&  \multicolumn{1}{c}{$E_{\xi}$ $\uparrow$}\\
		\hline
		w/o Geo&0.0052&0.8781&0.9791\\
		w/o Gran&0.0057&0.8708&0.9751\\
		w/o GIB&0.0053&0.8721&0.9777\\
		\textbf{Ours}&\textbf{0.0041}&\textbf{0.8808}&\textbf{0.9807}\\
		\hline
		
		\hline
	\end{tabular}
	\label{tab_dgc}
\end{table}

\begin{table}[t]
\small
	\renewcommand{\arraystretch}{0.9}
	\setlength\tabcolsep{8pt}
	\centering
	\caption{Ablation study on loss function.}
	\begin{tabular}{cccc}
		\hline
		\multicolumn{1}{c}{Settings}&  \multicolumn{1}{c}{$M$$\downarrow$}&\multicolumn{1}{c}{$F_{\beta}$$\uparrow$}&  \multicolumn{1}{c}{$E_{\xi}$$\uparrow$}\\
		\hline
		
		Bce&0.0099&0.6788&0.8625\\
		Fm&0.0057&0.8902&0.9785\\
		Iou&0.0087&0.8779&0.9765\\
		\textbf{Ours}&\textbf{0.0041}&\textbf{0.8808}&\textbf{0.9807}\\
		\hline
		
		\hline
	\end{tabular}
	\label{tab_loss}
\end{table}

\begin{table}[t]
\small
	\renewcommand{\arraystretch}{0.9}
	\setlength\tabcolsep{8pt}
	\centering
	\caption{Ablation study on unshuffle factor.}
	\begin{tabular}{cccc}
		\hline
		\multicolumn{1}{c}{Settings}&  \multicolumn{1}{c}{$M$$\downarrow$}&\multicolumn{1}{c}{$F_{\beta}$$\uparrow$}&  \multicolumn{1}{c}{$E_{\xi}$$\uparrow$}\\
		\hline
		
		1&0.0052&0.8642&0.9613\\
		1-2&0.0049&0.8713&0.9689\\
		1-2-4&0.0044&0.8795&0.9795\\
		\textbf{Ours}&\textbf{0.0041}&\textbf{0.8808}&\textbf{0.9807}\\
		\hline
		
		\hline
	\end{tabular}
	\label{tab_fac}
\end{table}

\begin{table}[t]
	\small
	\renewcommand{\arraystretch}{0.9}
	\setlength\tabcolsep{8pt}
	\centering
	\caption{Ablation study on different backbone.}
	\begin{tabular}{cccc}
		\hline
		\multicolumn{1}{c}{Settings}&  \multicolumn{1}{c}{$M$$\downarrow$}&\multicolumn{1}{c}{$F_{\beta}$$\uparrow$}&  \multicolumn{1}{c}{$E_{\xi}$$\uparrow$}\\
		\hline
		
		ResNet&0.0089&0.7911&0.9576\\
		Vgg&0.0125&0.7642&0.9015\\
		PVT&0.0063&0.8453&0.9683\\
		
		\textbf{Ours}&\textbf{0.0041}&\textbf{0.8808}&\textbf{0.9807}\\
		\hline
		
		\hline
	\end{tabular}
	\label{tab_back}
\end{table}
\subsection{Ablation Study}
 
To show the effectiveness of each component in our G$^2$HFNet (\emph{i.e.}, MDE, DGC, DSP and LGF), we conduct several ablation experiments, all numerical results are presented in  Table. \ref{tab_wo}.
\subsubsection{Effectiveness of different modules}
To evaluate the effectiveness of the modules proposed in this paper, we conduct three separate experiments, each involving the removal of one module. For instance, in experiment ``$w/o$ MDE", MDE module is removed. From  the numerical results of Table. \ref{tab_wo}, we can observe that the model's performance decreases as modules are removed, where $M$ gets improvements of $39.0\%$. Besides, compared to the ``$w/o$ DGC", Ours has decrease of $18.0\%$ in terms of $M$ and improvement of $1.3\%$  in terms of $F_{\beta}$ and $0.2\%$  in terms of $E_{\xi}$.  In addition, for the ``$w/o$ DSP", it can be observed from the Table. \ref{tab_wo} that the removal of DSP module leads to a decline in the model's detection performance. Furthermore, in the ``$w/o$ LGF", we replace the LGF module with $3\times3$ convolutional layer, and the comparison results indicate that the LGF module is effective in enhancing detection performance. Furthermore, we visualize the feature maps before and after each module to intuitively show their effectiveness. To make the comparison more quantitative, we additionally annotate each heatmap with an energy concentration score at the bottom-right corner. This metric reflects how well the activation responses are concentrated within the true target regions rather than being dispersed across the background, thus serving as a clear indicator of target-focused attention and activation compactness. As illustrated in Fig. \ref{fig_hot}, the MDE module effectively amplifies salient object regions, while the DSP module further sharpens object localization, producing more compact and well-defined activations. Correspondingly, the energy concentration values consistently increase after each module, demonstrating that the proposed designs progressively strengthen the model's ability to focus on meaningful target areas. These results collectively confirm the effectiveness of our module designs.

\subsubsection{Effectiveness of each component in MDE module}
To verify  the effectiveness of each component in MDE module, we conduct two experiments, as shown in Table. \ref{tab_mde}. For the ``$w/o$ U", we directly remove the four U-Net structures in the MDE module and feed the input features directly into the pyramid spatial attention and pyramid channel attention blocks. The features from these two attention blocks are then concatenated and fused to form the output of the MDE module. Besides, for the ``$w/o$ PA",  we remove the pyramid spatial attention and pyramid channel attention blocks, and instead concatenate and fuse the output features from the four U-Net structures.
\subsubsection{Effectiveness of each component in DGC module}
To illustrate the effectiveness of the each component in DGC module, we conduct three experiments presented in Table. \ref{tab_dgc}. For the ``$w/o$ Geo", we remove the geometric branch , and since only a single branch is used, the geo-gran interaction block cannot be applied in this case. Therefore, the feature from granular branch is directly used as the output features of DGC module. Similarly, for the ``$w/o$ Gran", when granular branch is removed, the feature from geometric branch is directly used as the output feature of the DGC module. Moreover, to verify  the effectiveness of  the geo-gran interaction block, we conduct the ``$w/o$ GIB" where the geo-gran interaction block is replaced by the element-wise addition to integrate the features from both branches. From the Table. \ref{tab_dgc}, numerical results reveal the importance of three componentes in DGC module. Additionally, we present the heatmap results of the geometric branch and granular branch in Fig. \ref{fig_gg}. As shown in the Fig. \ref{fig_gg}, the granular branch mainly focuses on the contour information of the object but contains more background noise, whereas the geometric branch concentrates on the positional information of the object and exhibits much less background interference. These observations validate the effectiveness of the dual-branch structure adopted in the DGC module.

\subsubsection{Effectiveness of different loss function}

To verify the impact of different loss function on  our G$^2$HFNet, we substitute the hybrid loss employed in the model with three separate individual losses (\emph{i.e.,} Bce, Fm, and Iou), as shown in Table. \ref{tab_loss}. To be specific, compared to ``Bce", our model gains improvement of $29.7\%$  in terms of $F_{\beta}$ and $13.7\%$  in terms of $E_{\xi}$. Besides,  compared to ``Fm", our G$^2$HFNet obtains decrease of $28.0\%$ in term of and $0.2\%$ in term of $E_{\xi}$, respectively. 
\subsubsection{Effectiveness of different unshuffle factor}
To evaluate how different unshuffle factors affect the model's performance, we conduct three experiments, as shown in the Table \ref{tab_fac}. For experiment $``1"$, we set the unshuffle factor to 1, meaning no multi-scale transformation is applied.
For experiment $``1-2"$, we use unshuffle factors of 1 and 2, generating two scales-the original resolution and a 2$\times$ downsampled resolution. Similarly, for experiment $``1-2-4"$, we adopt unshuffle factors of 1, 2, and 4, producing three scales-the original resolution, 2$\times$ downsampling, and 4$\times$ downsampling. From the results in the Table \ref{tab_fac}, we can observe a clear trend: as the number of scales decreases, the model's detection performance drops accordingly. This demonstrates the effectiveness of multi-scale feature generation through different unshuffle factors.
\subsubsection{Effectiveness of different backbone}
To investigate how different backbone choices affect the performance of G$^2$HFNet, we conduct a series of replacement experiments using VGG-16 \cite{simonyan2014very}, ResNet-34 \cite{he2016deep}, and PVT \cite{wang2021pyramid}, as summarized in Table \ref{tab_back}. When substituting Swin Transformer with VGG-16 or ResNet-34, G$^2$HFNet experiences clear performance degradation, for example, the Swin-based model outperforms ResNet-34 by $11.3\%$ in $F_\beta$ and $2.4\%$ in $E_{\xi}$, and surpasses VGG-16 by $15.2\%$ in $F_\beta$ and $8.8\%$ in $E_{\xi}$. These results indicate that conventional CNN backbones struggle to provide the rich hierarchical cues required by ORSI-SOD. We further evaluate transformer-based alternatives by replacing the Swin Transformer with  PVT. As shown in Table \ref{tab_back}, Swin Transformer again demonstrates superior capability: compared to PVT, G$^2$HFNet achieves a $34.9\%$ reduction in $M$ and a $4.2\%$ gain in $F_\beta$. These consistent advantages stem from Swin Transformer's shifted-window attention and hierarchical design, which capture both localized details and broader contextual structures more effectively than PVT's computationally intensive global attention.



\section{Conclusion}
In this paper, we propose a GeoGran-Aware Hierarchical Feature Fusion Network (G$^2$HFNet) for salient object detection, which fully exploits geometric and granular cues in optical remote sensing images. Specifically, G$^2$HFNet adopts a Swin Transformer backbone to extract multi-level features and integrates four key modules: a multi-scale detail enhancement (MDE) module, a dual-branch geo-gran complementary (DGC) module, a deep semantic perception (DSP) module, and a local-global guidance fusion (LGF) module. The MDE module is integrated into low-level features to handle object scale variations and capture fine details using U-Net structures combined with pyramid spatial and channel attention blocks. The DGC module addresses mid-level features by employing a granular branch for fine details and a geometric branch for positional cues, which are then fused in the geometric-granular interaction block. For high-level features, a self-attention mechanism is used to further refine positional information. Finally, the LGF module replaces traditional convolutions to integrate multi-level features along local and global dimensions, producing high-quality detection maps.

\bibliographystyle{IEEEtran}
\bibliography{reference}
\vspace{-4\baselineskip}
\end{document}